\documentclass[runningheads]{llncs}
\usepackage{graphicx}
\usepackage{multirow}
\usepackage[utf8]{inputenc} 
\usepackage[T1]{fontenc}    
\usepackage{hyperref}       
\usepackage{url}            
\usepackage{booktabs}       
\usepackage{amsfonts}       
\usepackage{nicefrac}       
\usepackage{microtype}      
\usepackage[table]{xcolor}        
\usepackage{subcaption}

\usepackage{amsmath}
\usepackage{algorithm}
\usepackage{algpseudocode}
\begin{document}
\title{From Verdict to Process: Agentic Reinforcement Learning for Multi-Stage Fact Verification}

\titlerunning{Agentic RL for Multi-Stage Fact Verification}


\author{
Rongxin Yang \and
Shenghong He \and
Siyuan Zhu \and
Chao Yu\thanks{Corresponding author.}
}
\authorrunning{R. Yang et al.}
\institute{
School of Computer Science and Engineering, Sun Yat-sen University, Guangzhou 510006, China
\\
\email{yangrx28@mail2.sysu.edu.cn, yuchao3@mail.sysu.edu.cn}
}

\maketitle              
\begin{abstract}
Recent approaches combining Large Language Models (LLMs) with retrieval-augmented reasoning have shown promise for automated fact verification. 
To process complex claims, these verification pipelines typically execute multi-stage workflows that coordinate tightly coupled modules, including claim decomposition, evidence gathering, and verdict prediction. 
However, existing methods optimize individual stages in isolation or rely on fixed heuristics, which limits adaptive coordination among stages and can lead to suboptimal outcomes.
In this work, we propose ProFact, an
agentic reinforcement learning framework for end-to-end optimization of
multi-stage fact verification trajectories. ProFact trains a unified policy to
coordinate claim decomposition, evidence seeking, answer generation, and
verdict prediction. 
To address the sparse and delayed supervision provided
by final veracity labels, ProFact introduces process-aware rewards that
provide stage-level learning signals throughout the verification process.
Empirical evaluation shows that ProFact consistently outperforms strong baselines in both verification performance and inference efficiency. These results highlight the effectiveness of process-aware trajectory optimization for multi-stage fact verification.

\end{abstract}
\section{Introduction}

\begin{figure*}
    \centering
    \includegraphics[width=1.0\linewidth]{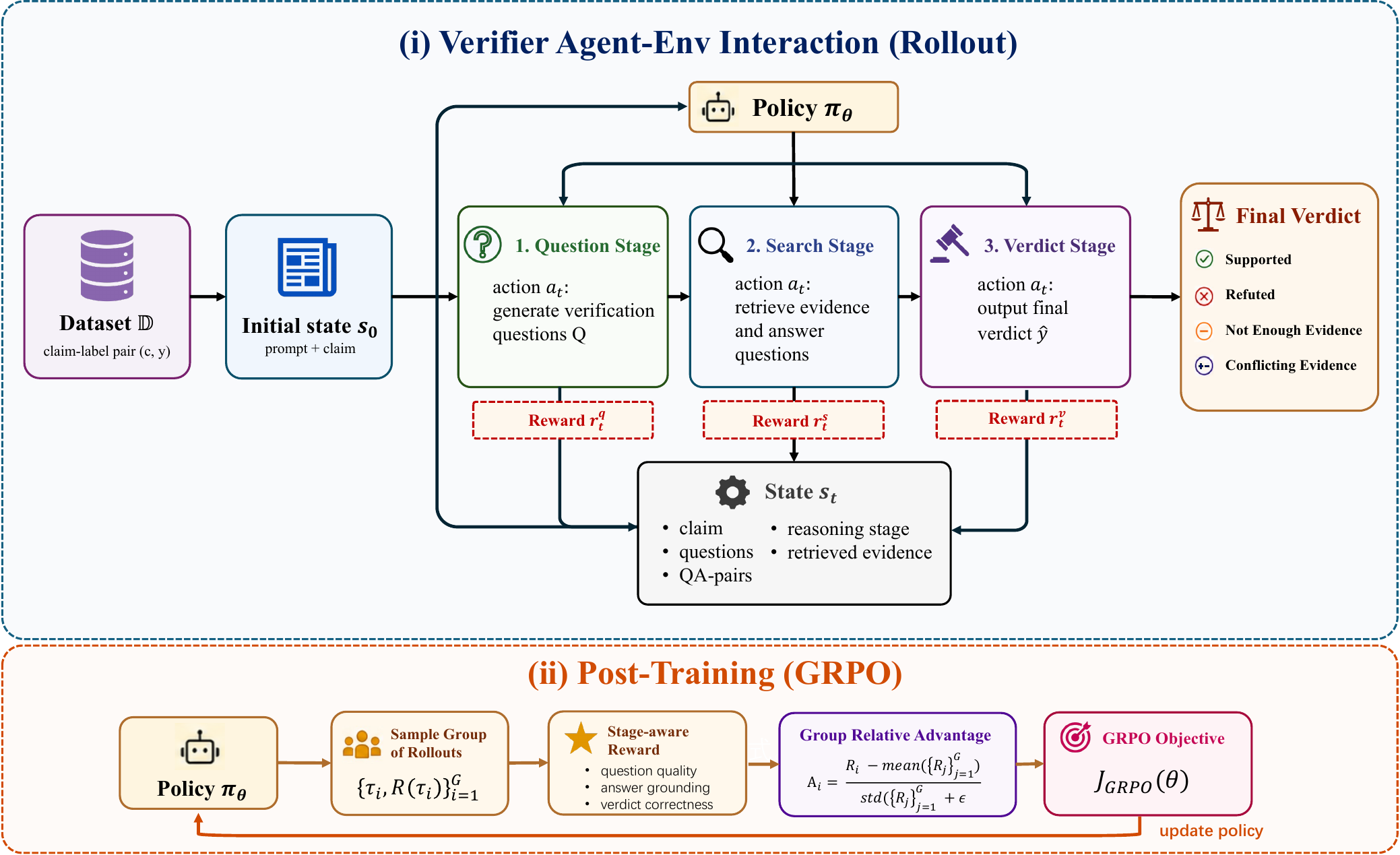}
    \caption{\textbf{Overview of ProFact.}
ProFact verifies claims through a multi-stage evidence-grounded trajectory involving claim decomposition, evidence retrieval, and veracity prediction, and optimizes this trajectory end-to-end with reinforcement learning.
}
    \label{fig:overview}
\end{figure*}

Automated fact verification aims to determine the veracity of natural-language claims by collecting and reasoning over reliable evidence\cite{guo2022survey}. This task is especially challenging in open-domain settings, where claims may involve multiple entities, implicit temporal constraints, and evidence scattered across heterogeneous sources. Rather than being solved as a single-step classification problem, real-world fact verification is inherently a multi-stage decision process that requires
identifying what should be checked, gathering relevant evidence, and aggregating
that evidence into a final verdict \cite{thorne2018fact,augenstein2019multifc,aly2021fact}.

Large Language Models (LLMs) have recently shown strong potential for fact verification when combined with retrieval-augmented reasoning\cite{lewis2020retrieval,yao2022react,vykopal2024generative}. By decomposing complex claims into verification questions and grounding reasoning in external evidence, LLM-based systems can produce more transparent and evidence-aware verification processes\cite{zhang2023towards}. These systems typically separate verification into multiple stages, such as claim decomposition, evidence retrieval, evidence-grounded answer generation, and verdict prediction. Recent methods such as InFact~\cite{rothermel2024infact}, HerO~\cite{yoon2024hero}, and DebateCV~\cite{he2026debating} further demonstrate the effectiveness of structured workflows for open-domain verification. 

However, existing workflow-based methods typically optimize these stages in isolation or rely on fixed heuristics to coordinate them. As a result, intermediate decisions may be poorly aligned with the end-to-end verification objective. For example, a decomposition strategy that produces plausible verification questions may not yield evidence that is most useful for final verdict prediction, and a retrieval strategy optimized independently may not best support downstream reasoning. This separation limits adaptive coordination across stages and can lead to suboptimal verification outcomes. 

To address this limitation, we propose \textbf{ProFact}, an agentic reinforcement learning framework for trajectory-level optimization of multi-stage fact verification. ProFact formulates verification as a long-horizon decision-making process, where a unified policy generates verification questions, produces retrieval actions for evidence seeking, generates evidence-grounded answers, and predicts the final veracity label. Instead of optimizing these behaviors separately, ProFact enables end-to-end optimization of the complete verification trajectory. Rather than treating these behaviors as independent modules, ProFact jointly optimizes them as a single trajectory, enabling earlier decisions (e.g., decomposition) to adapt to the requirements of later decisions (e.g., verdict prediction). Through exploration and exploitation, the policy can attempt different claim decomposition strategies, formulate varied search queries, and discover which evidence-seeking paths actually lead to correct verdicts. 
  
In practice, optimizing complete trajectories introduces a fundamental credit-assignment challenge. Supervision from the final veracity label is delayed and sparse: it indicates whether the overall prediction is correct, but does not reveal which intermediate decisions are responsible for the outcome. To provide more informative learning signals, ProFact introduces a process-aware reward function that supplies dense intermediate feedback for claim decomposition, evidence-grounded answer generation, and final verdict prediction. This reward design transforms a sparse outcome signal into stage-level supervision, allowing the policy to identify which intermediate behaviors contribute to or detract from final accuracy. Figure~\ref{fig:overview} provides an overview of the proposed framework. 

Our contributions are as follows:
\begin{itemize}
    \setlength{\itemsep}{0pt}
    \setlength{\parskip}{0pt}
    \setlength{\parsep}{0pt}
    \setlength{\topsep}{0pt}
    \item \textbf{Agentic Verification Framework.} We propose ProFact, which formulates
    open-domain fact verification as a long-horizon agentic decision-making
    problem and unifies claim decomposition, evidence retrieval, and verdict
    prediction within a single policy with explicit retrieval actions.
    \item \textbf{Process-Aware Trajectory Optimization.} We introduce a process-aware reward function that provides dense intermediate learning signals at each verification stage and enables end-to-end optimization of the complete verification trajectory via group-relative policy optimization.
    \item \textbf{Empirical Evaluation.} We conduct comprehensive experiments across four open-source backbones, demonstrating that ProFact improves verification performance and inference efficiency over strong baselines.
\end{itemize}

\section{Related Work}

\paragraph{\textbf{Fact Verification.}}
Automated fact verification aims to determine whether a claim is supported by reliable evidence. With the emergence of LLMs, several studies  \cite{khaliq2024ragar,zhang2023towards} have explored prompt-based verification methods, such as Chain-of-Thought reasoning \cite{wei2022chain}, to improve evidence analysis and verdict prediction. 
Recent LLM-based systems decompose fact verification into structured multi-stage pipelines. InFact~\cite{rothermel2024infact} achieves strong results by decomposing claim verification into a static, six-stage pipeline with engineered prompts and evidence retrieval, but its best performance relies on a proprietary backbone.
HerO~\cite{yoon2024hero} enhances evidence retrieval by generating hypothetical fact-checking documents, while employing separate LLM-based components for question generation and veracity prediction. DebateCV \cite{he2026debating} improves evidence scrutiny through multi-agent debate, with retrieval guided by strong proprietary models and fine-tuned models mainly used for final judgment. All of these methods are composed of independently trained or prompted modules, which limits adaptive coordination across stages. In contrast, ProFact formulates fact verification as a long-horizon agentic decision-making problem, where a unified policy is trained end-to-end to jointly optimize multi-stage verification behaviors.

\paragraph{\textbf{RL for LLM Agents.}}
RL has become a central paradigm for LLM post-training, evolving from preference-alignment methods such as RLHF \cite{ouyang2022traininglanguagemodelsfollow} to reasoning-oriented training with verifiable task rewards \cite{guo2025deepseek}.
 GRPO \cite{shao2024deepseekmathpushinglimitsmathematical} improves RL efficiency by estimating relative advantages from groups of sampled responses without requiring a separate critic. Beyond static reasoning tasks, recent work has extended RL to agentic LLMs that perform multi-step interaction, tool use, and reasoning under delayed feedback. RAGEN \cite{wang2025ragenunderstandingselfevolutionllm} provides the StarPO framework for analyzing and stabilizing multi-turn RL training of LLM agents. GiGPO \cite{feng2025groupingrouppolicyoptimizationllm} improves long-horizon agent training by introducing hierarchical group-based advantage estimation for finer-grained credit assignment. Meanwhile, retrieval-augmented methods such as Search-R1 \cite{jin2025searchr1trainingllmsreason} and Search-P1 \cite{xia2026search} train agents to interleave reasoning and search with trajectory-level feedback. While these methods demonstrate the effectiveness of RL for reasoning and search-based agentic tasks, ProFact explores its potential for end-to-end optimization in multi-stage fact verification.

\section{Method}

\subsection{Task Definition}

We formulate fact verification as a finite-horizon Markov decision process (MDP), in which a verifier agent iteratively acquires evidence, updates its reasoning state, and ultimately produces a veracity verdict. Formally, the MDP is defined as $\mathcal{M}=(\mathcal{S},\mathcal{A},P,r,\rho_0,T)$, where $\mathcal{S}$ denotes the state space, $\mathcal{A}$ the action space, $P$ the transition dynamics, $r$ the reward function, $\rho_0$ the initial-state distribution generated from the claim dataset, and $T$ the maximum decision horizon.

Let $\mathcal{D}=\{(c_i, y_i)\}_{i=1}^{N}$ denote a dataset of claims, where each $c_i$ is a natural-language claim and $y_i \in \mathcal{Y}$ denotes its veracity label. In this work, we instantiate the formulation on AVeriTeC \cite{schlichtkrull2023averitec}, whose label space is given by
\[
\mathcal{Y}_{\text{AVeriTeC}}
=
\left\{
\begin{array}{l}
\textsc{Supported},\ \textsc{Not Enough Evidence},\\
\textsc{Refuted},\ \textsc{Conflicting Evidence}
\end{array}
\right\}.
\]

At the beginning of each episode, a claim-label pair $(c,y)\sim\mathcal{D}$ is sampled and used to initialize the state $s_0\sim\rho_0$. At each time step $t\in\{0,\dots,T-1\}$, the agent observes the current state $s_t$, generates an action $a_t$ according to the policy $\pi_\theta$, and receives a reward $r_t$. This process generates a trajectory $\tau=(s_0,a_0,r_0,\dots,s_{T-1},a_{T-1},r_{T-1})$, whose return is defined as $R(\tau)=\sum_{t=0}^{T-1} r_t$.

The learning objective is to maximize the expected return over claim-conditioned verification trajectories:
\begin{equation}
\label{eq:objective}
J(\theta)
=
\mathbb{E}_{(c,y)\sim\mathcal{D},\, \tau\sim p_\theta(\tau\mid c)}
\left[
R(\tau)
\right],
\end{equation}
where $p_\theta(\tau\mid c)$ denotes the trajectory distribution generated by the policy $\pi_\theta$.

\subsection{Overall Framework}

Our framework instantiates the MDP as a three-stage verification rollout that mirrors the workflow of human fact-checking, consisting of \textsc{Question}, \textsc{Search}, and \textsc{Verdict} stages. Given a claim, the agent first generates verification questions in the \textsc{Question} stage, then retrieves evidence and produces evidence-grounded answers in the \textsc{Search} stage, and finally predicts the veracity label in the \textsc{Verdict} stage. Algorithm~\ref{alg:averitec-rollout} summarizes the detailed rollout procedure.

\begin{figure*}
    \centering
    \includegraphics[width=0.8\linewidth]{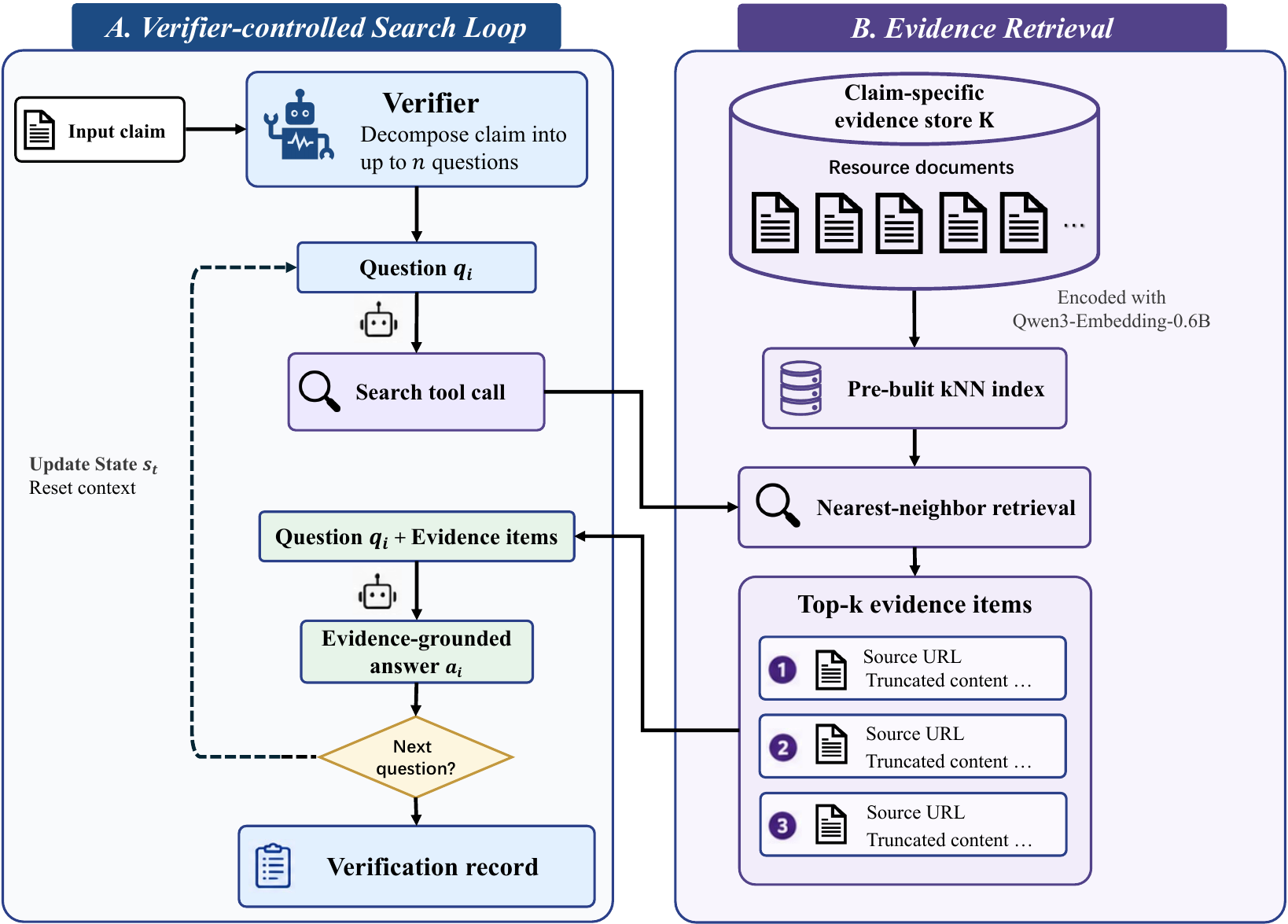}
    \caption{Overview of the \textsc{Search} stage.}
    \label{fig:search}
\end{figure*}

Among the three stages, the \textsc{Search} stage plays a central role because it directly determines the quality of the evidence available for downstream reasoning and final verdict prediction. As illustrated in Figure \ref{fig:search}, in the \textsc{Search} stage, the agent iteratively retrieves evidence for each verification question $q_i$ via explicit tool calling. 
Rather than relying on internal knowledge, the agent generates structured search actions to query the evidence store $\mathcal{K}$. 
To retrieve the most relevant context, we implement a semantic search mechanism: resource documents are encoded using Qwen3-Embedding-0.6B, and the agent's queries are matched against a pre-built kNN index to fetch the top-$k$ nearest items. 
Conditioned on this retrieved evidence, the agent synthesizes an evidence-grounded answer $a_i$. 
The completed $(q_i, a_i)$ pair is then appended to the global verification record, and the transient search context is reset before the agent advances to the next question $q_{i+1}$.

For post-training, we optimize the policy with GRPO over grouped multi-stage rollouts. 
For each claim $c_i$, we sample a group of $G$ trajectories 
$\{\tau_{i,g}\}_{g=1}^{G}$ using the old policy 
$\pi_{\theta_{\mathrm{old}}}$. 
Each trajectory is scored by the process-aware reward function, and the resulting 
trajectory-level returns are normalized within the group to compute group-relative 
advantages.

Specifically, the advantage of the $g$-th trajectory for claim $c_i$ is computed as
\begin{equation}
\label{eq:grpo_advantage}
\hat{A}_{i,g}
=
\frac{
R(\tau_{i,g})-\mu_i
}{
\sigma_i+\epsilon
},
\quad
\mu_i
=
\frac{1}{G}
\sum_{g=1}^{G}
R(\tau_{i,g}),
\quad
\sigma_i
=
\sqrt{
\frac{1}{G}
\sum_{g=1}^{G}
\left(
R(\tau_{i,g})-\mu_i
\right)^2
}.
\end{equation}
The policy is then updated using the following clipped GRPO objective~\cite{shao2024deepseekmathpushinglimitsmathematical}:
\begin{equation}
\label{eq:grpo}
\begin{aligned}
\mathcal{J}_{\mathrm{GRPO}}(\theta)
=
\mathbb{E}
\left[
\frac{1}{G}\sum_{i=1}^{G}
\frac{1}{|\tau_{i,g}|}
\sum_{\ell=1}^{|\tau_{i,g}|}
\left(
\min
\left(
\rho_{i,g,\ell}\hat{A}_{i,g},
\bar{\rho}_{i,g,\ell}\hat{A}_{i,g}
\right)
-
\beta D_{\mathrm{KL}}^{i,g,\ell}
\right)
\right],
\end{aligned}
\end{equation}
where the token-level policy ratio and its clipped version are defined as
\begin{equation}
\label{eq:grpo_ratio}
\rho_{i,g,\ell}
=
\frac{
\pi_{\theta}(o_{i,g,\ell}\mid h_{i,g,\ell})
}{
\pi_{\theta_{\mathrm{old}}}(o_{i,g,\ell}\mid h_{i,g,\ell})
},
\quad
\bar{\rho}_{i,g,\ell}
=
\operatorname{clip}
\left(
\rho_{i,g,\ell},1-\epsilon,1+\epsilon
\right).
\end{equation}
Here, $i$ indexes the claim, $g$ indexes the sampled trajectory within the group, 
and $\ell$ indexes the token position in trajectory $\tau_{i,g}$. 
The term $o_{i,g,\ell}$ denotes the $\ell$-th generated token, 
$h_{i,g,\ell}$ denotes the generation history before producing this token, and 
$|\tau_{i,g}|$ denotes the number of generated tokens in the trajectory. 
$R(\tau_{i,g})$ is the total process-aware return of the trajectory, and 
$\hat{A}_{i,g}$ is broadcast to all generated tokens in the trajectory during the 
policy update. 
The coefficient $\epsilon$ controls the clipping range, 
$\epsilon_{\mathrm{std}}$ is a small constant for numerical stability, and 
$\beta$ controls the strength of KL regularization. 
The term $D_{\mathrm{KL}}^{i,g,\ell}$ penalizes the deviation of the updated policy 
from the reference policy $\pi_{\mathrm{ref}}$ at token position $\ell$. 
This objective encourages the policy to assign higher likelihood to trajectories 
with better question decomposition, stronger evidence use, and more reliable 
final verdicts, while constraining the updated policy from drifting too far from 
the reference model.

\begin{algorithm}[t]
\footnotesize
\caption{Multi-stage Rollout for Fact Verification}
\label{alg:averitec-rollout}
\begin{algorithmic}[1]
\Require Claim-label pair $(c,y)$, policy $\pi_\theta$, evidence store $\mathcal{K}$, horizon $T$
\State $s_1\gets(c,\textsc{Question})$, $\tau\gets\emptyset$, $R(\tau)\gets0$
\For{$t=1$ to $T$}
    \State $a_t\sim\pi_\theta(\cdot\mid s_t)$ \Comment{Unified action generation}
    \If{$\mathrm{stage}(s_t)=\textsc{Question}$}
        \State $\mathcal{\hat Q}\gets\mathrm{ParseQ}(a_t)$
        \State $r_t\gets r^q(s_t,\mathcal{\hat Q};\mathcal{Q^*})$
        \State $q \gets \mathcal{\hat Q}[0]$ \Comment{Get the first question}
        \State $s_{t+1}\gets(c,\textsc{Search},\mathcal{\hat Q},q,\emptyset)$ \Comment{Init with empty history and evidence}
    \ElsIf{$\mathrm{stage}(s_t)=\textsc{Search}$}
        \If{$\mathrm{IsToolCall}(a_t)$} \Comment{Check if action is a tool call}
            \State $e \gets\mathrm{Retrieve}(a_t, \mathcal{K})$
            \State $s_{t+1}\gets(c,\textsc{Search},\mathcal{\hat Q},\mathcal{\hat H},q, e)$ \Comment{Update evidence in state}
        \ElsIf{$\mathrm{IsAnswer}(a_t)$} \Comment{Action is an answer}
            \State $\mathcal{\hat H}\gets\mathcal{\hat H}\cup\{(q, a_t)\}$ \Comment{Current question being answered}
            \If{$|\mathcal{\hat H}| = |\mathcal{\hat Q}|$} \Comment{Check if all questions are answered}
                \State $r_t\gets r^s(s_t,\mathcal{\hat H};\mathcal{H^*})$
                \State $s_{t+1}\gets(c,\textsc{Verdict},\mathcal{\hat Q},\mathcal{\hat H})$
            \Else
                \State $q \gets \mathcal{\hat Q}[|\mathcal{\hat H}|]$ \Comment{Get the next question}
                \State $s_{t+1}\gets(c,\textsc{Search},\mathcal{\hat Q},\mathcal{\hat H},q,\emptyset)$ \Comment{Move to next question}
            \EndIf
        \EndIf
    \ElsIf{$\mathrm{stage}(s_t)=\textsc{Verdict}$}
        \State $\hat{y}\gets\mathrm{ParseY}(a_t)$
        \State $r_t\gets r^v(s_t,\hat{y};y)$
        \State $s_{t+1}\gets\textsc{End}$
    \EndIf
    \State $\tau\gets\mathrm{Append}(\tau,(s_t,a_t,r_t))$
    \State $R(\tau)\gets R(\tau)+r_t$
\EndFor
\State \Return $\tau$, $R(\tau)$
\end{algorithmic}
\vspace{-0.5em}
\end{algorithm}

\subsection{Reward Function}

We design a process-aware reward function that provides intermediate supervision for different verification behaviors. For text-based intermediate outputs, we use a METEOR-based matching score against the gold annotations. Given a predicted set $X=\{x_i\}_{i=1}^{m}$ and a gold set $Y=\{y_j\}_{j=1}^{n}$, we compute pairwise METEOR similarities and obtain the best one-to-one alignment via maximum-weight bipartite matching:
\begin{equation}
\label{eq:match_score}
\mathrm{Match}(X,Y)
=
\frac{1}{|Y|}
\max_{\mathcal{M}}
\sum_{(i,j)\in \mathcal{M}}
\mathrm{METEOR}(x_i,y_j),
\end{equation}
where $\mathcal{M}$ denotes a valid one-to-one assignment between predicted and gold items. In the \textsc{Question} stage, $X$ and $Y$ correspond to the generated and gold verification questions, respectively, so the reward evaluates the quality of claim decomposition. In the \textsc{Search} stage, each question-answer pair is converted into a comparison string by concatenating the question and answer, and the same matching score is used to evaluate evidence-grounded answer synthesis. In the \textsc{Verdict} stage, the reward is based on whether the predicted veracity label matches the gold label. The overall process-aware reward is therefore defined as:
\begin{equation}
\label{eq:stage_reward}
r_t =
\begin{cases}
\mathrm{Match}(\hat{\mathcal{Q}}, \mathcal{Q}^{*}),
& \mathrm{stage}(s_t)=\textsc{Question},\\
\mathrm{Match}(\hat{\mathcal{H}}, \mathcal{H}^{*}),
& \mathrm{stage}(s_t)=\textsc{Search},\\
\mathbb{I}[\hat{y}=y],
& \mathrm{stage}(s_t)=\textsc{Verdict},
\end{cases}
\end{equation}
where $\hat{\mathcal{Q}}$ and $\mathcal{Q}^{*}$ denote predicted and gold question sets, $\hat{\mathcal{H}}$ and $\mathcal{H}^{*}$ denote predicted and gold question-answer sets, and $\mathbb{I}[\cdot]$ is the indicator function. Invalid or empty outputs are assigned zero reward. This reward design provides denser supervision than a purely terminal signal, encouraging the policy to improve claim decomposition, evidence-grounded answer generation, and final veracity prediction throughout the trajectory.

\section{Experiments}

\subsection{Experimental Setup}

\paragraph{\textbf{Dataset.}}
We conduct experiments on AVeriTeC \cite{schlichtkrull2023averitec}, a benchmark designed for real-world fact verification with open-world evidence acquisition. Each instance consists of a natural-language claim, a veracity label, and annotated question--answer evidence used to support the verification process. AVeriTeC also provides a static knowledge store of pre-collected web documents, which we use as the evidence source for retrieval during the \textsc{Search} stage. This setup enables reproducible evaluation while preserving the open-world nature of evidence-seeking fact verification.

\paragraph{\textbf{Evaluation Metrics.}}
We report four evaluation metrics that cover both intermediate evidence construction and final verification performance.
Q-only METEOR evaluates the generated verification questions against the gold questions, reflecting the model's ability to identify relevant information needs for claim verification.
Q\&A METEOR evaluates the generated question--answer pairs by measuring their alignment with the annotated verification process.
Accuracy measures final veracity classification performance.
Following the official AVeriTeC protocol, AVeriTeC Score measures end-to-end verification performance: a prediction is considered correct only if the evidence score exceeds the cutoff $\lambda=0.25$ and the predicted veracity label is correct.
This metric therefore jointly captures evidence sufficiency and final decision accuracy.

\paragraph{\textbf{Baselines.}}
We compare ProFact with three representative baselines. \textbf{Consistency} is a prompting-based baseline that uses the released baseline questions-answer pairs as fixed evidence input, and applies repeated generation with consistency-based aggregation to predict the final veracity label. \textbf{InFact} \cite{rothermel2024infact} is a workflow-based fact-checking method that performs multi-step evidence acquisition and reasoning. \textbf{HerO} \cite{yoon2024hero} is a strong fact-verification system that enhances evidence retrieval with hypothetical fact-checking documents and uses fine-tuned LLMs for veracity prediction. To ensure a fair comparison, we evaluate all methods under the same evidence source and report results across 4 open-source backbone models: Qwen2.5-3B-Instruct, Qwen2.5-7B-Instruct, Qwen3-4B-Instruct-2507, and Qwen3-8B. For brevity, we refer to these backbones as Qwen2.5-3B, Qwen2.5-7B, Qwen3-4B, and Qwen3-8B throughout the rest of this paper. For Qwen3-8B, we use the default thinking mode.

\begin{table}[t]
\centering
\setlength{\tabcolsep}{8pt}
\caption{Main results of different methods on AVeriTeC. 
\emph{w/o PR} removes the process rewards assigned to intermediate stages and retains only the outcome reward for training.}
\label{tab:main_results}
\begin{tabular}{llcccc}
\toprule
\multirow{2}{*}[-0.2em]{Method} 
& \multirow{2}{*}[-0.2em]{LLMs} 
& \multicolumn{2}{c}{METEOR} 
& \multirow{2}{*}[-0.2em]{Accuracy} 
& \multirow{2}{*}[-0.25em]{\shortstack{AVeriTeC\\Score}} \\
\cmidrule(lr){3-4}
& & Q-only & Q\&A & & \\
\midrule
\multirow{4}{*}{Consistency}
& Qwen2.5-3B & 29.65 & 23.86 & 29.60 & 15.60 \\
& Qwen2.5-7B  & 29.65 & 23.86 & 41.00 & 20.60 \\
& Qwen3-4B    & 29.65 & 23.86 & 51.00 & 21.40 \\
& Qwen3-8B    & 29.65 & 23.86 & 45.60 & 22.40 \\
\midrule
\multirow{4}{*}{InFact}
& Qwen2.5-3B & 37.23 & 27.99 & 14.83 & 8.54 \\
& Qwen2.5-7B  & 33.74 & 26.79 & 37.72 & 23.71 \\
& Qwen3-4B    & 40.09 & 29.12 & 56.91 & \underline{45.29} \\
& Qwen3-8B    & 35.54 & 26.27 & 55.07 & 31.50 \\
\midrule
\multirow{4}{*}{HerO}
& Qwen2.5-3B & 44.29 & 30.54 & 61.00 & \underline{43.40} \\
& Qwen2.5-7B  & 44.29 & 30.54 & \underline{67.80} & \underline{42.80} \\
& Qwen3-4B    & 44.29 & 30.54 & \underline{68.20} & 43.60 \\
& Qwen3-8B    & 44.29 & 30.54 & \underline{69.20} & \underline{44.40} \\
\midrule
\multirow{4}{*}{\textbf{ProFact}}
& Qwen2.5-3B  & 46.01 & 31.14 & \textbf{68.80} & \textbf{47.80} \\
& Qwen2.5-7B  & 45.26 & 30.36 & \textbf{70.20} & \textbf{48.00} \\
& Qwen3-4B    & 46.08 & 30.11 & \textbf{69.60} & \textbf{46.20} \\
& Qwen3-8B    & 46.05 & 30.02 & \textbf{70.28} & \textbf{46.40} \\
\midrule
\multirow{4}{*}{\hspace{1em}w/o PR}
& Qwen2.5-3B  & 38.98 & 27.20 & \underline{64.60} & 34.40 \\
& Qwen2.5-7B  & 33.79 & 25.67 & 62.80 & 31.00 \\
& Qwen3-4B    & 39.23 & 27.88 & 66.00 & 39.00 \\
& Qwen3-8B    & 40.29 & 27.26 & 64.20 & 35.40 \\
\bottomrule
\end{tabular}
\end{table}

\paragraph{\textbf{Implementation Details.}}
We initialize the verifier from the corresponding open-source backbone and perform post-training on the training split of AVeriTeC. All reported results are evaluated on the development set. For each claim, we sample 8 trajectories to compute group-relative advantages. For batching, we set the mini-batch size to 32 and the micro-batch size to 4.
 We enable a KL regularization term with coefficient 0.001. The verifier operates in a three-stage pipeline, with at most 12 interaction steps per episode. During evaluation, we use deterministic decoding with temperature set to $0$, limit the verifier to at most five generated verification questions, and keep the evidence store fixed across all methods. For retrieval, each search call returns the top 3 evidence items, and we use the same retrieval budget for all compared systems. To make evidence-grounded outputs comparable to the gold annotations and avoid advantages from overly long reasoning traces, we constrain the maximum number of generated verification questions to five for all methods.

\subsection{Main Results}

The main results comparing ProFact with the baseline methods are presented in Table~\ref{tab:main_results}.  
For Consistency and HerO, the Q-only and Q\&A scores remain unchanged across backbone models because the intermediate question-answer evidence is either fixed or produced by separate retrieval/question-generation modules, whereas the evaluated backbone is used only for final veracity prediction. 
Overall, ProFact achieves the best verification performance across the evaluated backbone models, consistently outperforming the strongest baseline in terms of the AVeriTeC Score. 
This demonstrates that optimizing the verifier over multi-stage trajectories improves the overall effectiveness of evidence-grounded fact verification.

ProFact also achieves the highest Accuracy across all backbones, suggesting that the proposed training strategy improves final label prediction in addition to intermediate evidence construction. 
For intermediate reasoning quality, ProFact obtains the best Q-only METEOR score, indicating stronger claim decomposition. 
On Q\&A METEOR, ProFact remains competitive with the strongest baselines, while achieving better verification performance. 
These results suggest that ProFact does not merely optimize textual overlap with gold evidence, but learns verification trajectories that better support reliable final decision making.

It is worth noting that larger backbones do not necessarily achieve higher Accuracy or AVeriTeC Score, which is consistent with findings on inverse scaling \cite{mckenzie2023inverse}. 
In the context of fact verification, larger models may exhibit greater reliance on parametric knowledge, learned plausibility priors, or shortcut patterns acquired during pretraining. 
Although these priors can support fluent reasoning, they may also bias the model away from the retrieved evidence when the verification decision requires fine-grained evidence assessment.
Consequently, models with larger parameter counts may not consistently yield more reliable evidence-grounded verification, leading to non-monotonic trends across evaluation metrics.

\subsection{Ablation Study}

We study the effect of process rewards by comparing ProFact with w/o PR, where PR denotes the intermediate process rewards used in the \textsc{Question} and \textsc{Search} stages. This variant removes these rewards and retains only the verdict reward. As shown in Table~\ref{tab:main_results}, removing process rewards consistently degrades performance across all backbone models, particularly in intermediate evidence quality and the overall AVeriTeC Score. This indicates that final-label supervision alone is too sparse and delayed to reliably assign credit to the intermediate decisions that shape the verification trajectory. By providing denser stage-level feedback, process rewards help the model distinguish whether verification success or failure stems from claim decomposition, evidence retrieval, or evidence-grounded reasoning. These results demonstrate that process-aware rewards are important for improving credit assignment in multi-stage fact verification.

\begin{table}[t]
\centering
\setlength{\tabcolsep}{7pt}
\caption{Efficiency comparison of different methods.}
\label{tab:efficiency}
\begin{tabular}{llccc}
\toprule
\multirow{2}{*}[-0.2em]{Method} 
& \multirow{2}{*}[-0.2em]{LLMs} 
& \multirow{2}{*}[-0.2em]{\shortstack{Time per\\Claim (s)}} 
& \multicolumn{2}{c}{Total Tokens} \\
\cmidrule(lr){4-5}
& & & Input & Output \\
\midrule
\multirow{4}{*}{InFact}
& Qwen2.5-3B & 16.32 & 55.15M & 0.54M \\
& Qwen2.5-7B & 50.44 & 131.44M & 0.66M \\
& Qwen3-4B   & 114.81 & 52.40M & 3.65M \\
& Qwen3-8B   & 288.00 & 34.04M & 12.63M \\
\midrule
\multirow{4}{*}{Ours}
& Qwen2.5-3B & 7.29 & 7.26M & 0.38M \\
& Qwen2.5-7B & 7.82 & 7.58M & 0.36M \\
& Qwen3-4B   & 7.81 & 7.22M & 0.43M \\
& Qwen3-8B   & 22.06 & 6.59M & 2.01M \\
\bottomrule
\end{tabular}
\end{table}

\section{Analysis}

\subsection{Efficiency Analysis}

Table~\ref{tab:efficiency} analyzes the resource consumption and inference efficiency of the verification process. The token cost reported in the table is computed as the total input and output token consumption over the 500 claims in the development set. We compare ProFact with InFact since both methods rely on a single backbone to execute the complete verification workflow end-to-end, making their inference costs comparable. While achieving stronger verdict performance, ProFact consistently reduces cost across all backbone models, requiring less time and substantially fewer tokens. Compared with InFact \cite{rothermel2024infact}, ProFact streamlines the verification workflow by removing claim rewriting and merging verdict prediction with justification generation, thereby reducing redundant intermediate steps. In addition, ProFact uses context isolation as an auxiliary mechanism to avoid unnecessary context expansion across decomposed questions. Through training, it further learns a less redundant and more targeted verification strategy, reducing unnecessary evidence-seeking while preserving effective reasoning.

\begin{table}[t]
\centering
\setlength{\tabcolsep}{7pt}
\caption{Comparison of different RL algorithms.}
\label{tab:rl_compare}
\begin{tabular}{llcccc}
\toprule
\multirow{2}{*}[-0.2em]{Algorithm} 
& \multirow{2}{*}[-0.2em]{LLMs}
& \multicolumn{2}{c}{METEOR} 
& \multirow{2}{*}[-0.2em]{\shortstack{Accuracy}} 
& \multirow{2}{*}[-0.2em]{\shortstack{AVeriTeC\\Score}} \\
\cmidrule(lr){3-4}
& & Q-only & Q\&A & & \\
\midrule
\multirow{2}{*}{PPO}
& Qwen2.5-3B
& 41.63 
& 28.21 
& 65.46 
& 31.00 \\
& Qwen2.5-7B
& 37.18
& 27.23
& 67.74
& 37.27 \\
\midrule
\multirow{2}{*}{GRPO}
& Qwen2.5-3B
& 46.01 
& 31.14 
& \textbf{68.80} 
& \textbf{47.80} \\
& Qwen2.5-7B
& 45.26
& 30.36
& \textbf{70.20}
& \textbf{48.00} \\
\midrule
\multirow{2}{*}{DAPO}
& Qwen2.5-3B
& 44.59 
& 29.71
& 66.40 
& 42.00 \\
& Qwen2.5-7B
& 43.36
& 29.32
& 68.80
& 42.80 \\
\midrule
\multirow{2}{*}{GiGPO}
& Qwen2.5-3B
& 46.22
& 29.23 
& 66.00 
& 38.80 \\
& Qwen2.5-7B
& 44.49
& 29.69
& 69.60
& 43.20 \\
\bottomrule
\end{tabular}
\end{table}

\subsection{Effect of RL Algorithms}

We further study how different RL optimization algorithms affect ProFact. As shown in Table~\ref{tab:rl_compare}, GRPO delivers the strongest overall results among the compared algorithms. This indicates that GRPO is well suited to multi-stage fact verification, as its group-relative objective compares multiple rollouts for the same claim and provides a stable trajectory-level learning signal. PPO, by contrast, requires learning a value function for state-level return estimation, which is challenging in ProFact due to heterogeneous stage rewards and retrieval-dependent outcomes.
 GiGPO improves credit assignment through anchor-state grouping, but verification tasks typically contain few reusable or recurring anchor states, limiting the benefit of this finer-grained grouping mechanism. These results support the use of GRPO as the default optimization algorithm in ProFact.

%
%

\section{Conclusion}

In this paper, we propose ProFact, an agentic reinforcement learning framework for multi-stage fact verification. ProFact formulates verification as a finite-horizon decision process and optimizes evidence-seeking trajectories with process-aware rewards. Experiments on AVeriTeC demonstrate that ProFact consistently improves verification effectiveness while reducing inference cost. Further analyses confirm that intermediate process supervision is critical for learning effective verification trajectories and that trajectory-level reinforcement learning provides a suitable optimization signal for long-horizon verification. These results highlight the effectiveness of process-aware trajectory optimization for multi-stage fact verification.


%
%
%
\bibliographystyle{splncs04}
\bibliography{ref}

@misc{guo2022survey,
  title={A survey on automated fact-checking},
  author={Guo, Zhijiang and Schlichtkrull, Michael and Vlachos, Andreas},
  journal={Transactions of the association for computational linguistics},
  volume={10},
  pages={178--206},
  year={2022}
}

@misc{thorne2018fact,
  title={The fact extraction and VERification (FEVER) shared task},
  author={Thorne, James and Vlachos, Andreas and Cocarascu, Oana and Christodoulopoulos, Christos and Mittal, Arpit},
  booktitle={Proceedings of the first workshop on Fact Extraction and VERification (FEVER)},
  pages={1--9},
  year={2018}
}

@misc{aly2021fact,
  title={The fact extraction and VERification over unstructured and structured information (FEVEROUS) shared task},
  author={Aly, Rami and Guo, Zhijiang and Schlichtkrull, Michael and Thorne, James and Vlachos, Andreas and Christodoulopoulos, Christos and Cocarascu, Oana and Mittal, Arpit},
  booktitle={Proceedings of the Fourth Workshop on Fact Extraction and VERification (FEVER)},
  pages={1--13},
  year={2021}
}

@misc{wei2022chain,
  title={Chain-of-thought prompting elicits reasoning in large language models},
  author={Wei, Jason and Wang, Xuezhi and Schuurmans, Dale and Bosma, Maarten and Xia, Fei and Chi, Ed and Le, Quoc V and Zhou, Denny and others},
  journal={Advances in neural information processing systems},
  volume={35},
  pages={24824--24837},
  year={2022}
}

@misc{zhang2023towards,
  title={Towards llm-based fact verification on news claims with a hierarchical step-by-step prompting method},
  author={Zhang, Xuan and Gao, Wei},
  booktitle={Proceedings of the 13th international joint conference on natural language processing and the 3rd conference of the asia-pacific chapter of the association for computational linguistics (volume 1: Long papers)},
  pages={996--1011},
  year={2023}
}

@misc{khaliq2024ragar,
  title={Ragar, your falsehood radar: Rag-augmented reasoning for political fact-checking using multimodal large language models},
  author={Khaliq, Mohammed Abdul and Chang, Paul Yu-Chun and Ma, Mingyang and Pflugfelder, Bernhard and Mileti{\'c}, Filip},
  booktitle={Proceedings of the Seventh Fact Extraction and VERification Workshop (FEVER)},
  pages={280--296},
  year={2024}
}

@misc{schlichtkrull2023averitec,
  title={Averitec: A dataset for real-world claim verification with evidence from the web},
  author={Schlichtkrull, Michael and Guo, Zhijiang and Vlachos, Andreas},
  journal={Advances in Neural Information Processing Systems},
  volume={36},
  pages={65128--65167},
  year={2023}
}

@misc{augenstein2019multifc,
  title={MultiFC: A real-world multi-domain dataset for evidence-based fact checking of claims},
  author={Augenstein, Isabelle and Lioma, Christina and Wang, Dongsheng and Lima, Lucas Chaves and Hansen, Casper and Hansen, Christian and Simonsen, Jakob Grue},
  booktitle={Proceedings of the 2019 conference on empirical methods in natural language processing and the 9th international joint conference on natural language processing (EMNLP-IJCNLP)},
  pages={4685--4697},
  year={2019}
}

@misc{rothermel2024infact,
  title={InFact: A strong baseline for automated fact-checking},
  author={Rothermel, Mark and Braun, Tobias and Rohrbach, Marcus and Rohrbach, Anna},
  booktitle={Proceedings of the Seventh Fact Extraction and VERification Workshop (FEVER)},
  pages={108--112},
  year={2024}
}

@misc{yoon2024hero,
  title={HerO at AVeriTeC: The herd of open large language models for verifying real-world claims},
  author={Yoon, Yejun and Jung, Jaeyoon and Yoon, Seunghyun and Park, Kunwoo},
  booktitle={Proceedings of the Seventh Fact Extraction and VERification Workshop (FEVER)},
  pages={130--136},
  year={2024}
}

@misc{yao2022react,
  title={React: Synergizing reasoning and acting in language models},
  author={Yao, Shunyu and Zhao, Jeffrey and Yu, Dian and Du, Nan and Shafran, Izhak and Narasimhan, Karthik R and Cao, Yuan},
  booktitle={The eleventh international conference on learning representations},
  year={2022}
}

@misc{lewis2020retrieval,
  title={Retrieval-augmented generation for knowledge-intensive nlp tasks},
  author={Lewis, Patrick and Perez, Ethan and Piktus, Aleksandra and Petroni, Fabio and Karpukhin, Vladimir and Goyal, Naman and K{\"u}ttler, Heinrich and Lewis, Mike and Yih, Wen-tau and Rockt{\"a}schel, Tim and others},
  journal={Advances in neural information processing systems},
  volume={33},
  pages={9459--9474},
  year={2020}
}

@misc{vykopal2024generative,
  title={Generative large language models in automated fact-checking: A survey},
  author={Vykopal, Ivan and Pikuliak, Mat{\'u}{\v{s}} and Ostermann, Simon and {\v{S}}imko, Mari{\'a}n},
  journal={arXiv preprint arXiv:2407.02351},
  year={2024}}

@misc{he2026debating,
  title={Debating truth: Debate-driven claim verification with multiple large language model agents},
  author={He, Haorui and Li, Yupeng and Wen, Dacheng and Chen, Yang and Cheng, Reynold and Chen, Donglong and Lau, Francis CM},
  booktitle={Proceedings of the ACM Web Conference 2026},
  pages={8851--8861},
  year={2026}}

@misc{ouyang2022traininglanguagemodelsfollow,
      title={Training language models to follow instructions with human feedback}, 
      author={Long Ouyang and Jeff Wu and Xu Jiang and Diogo Almeida and Carroll L. Wainwright and Pamela Mishkin and Chong Zhang and Sandhini Agarwal and Katarina Slama and Alex Ray and John Schulman and Jacob Hilton and Fraser Kelton and Luke Miller and Maddie Simens and Amanda Askell and Peter Welinder and Paul Christiano and Jan Leike and Ryan Lowe},
      year={2022},
      eprint={2203.02155},
      archivePrefix={arXiv},
      primaryClass={cs.CL},
      url={https://arxiv.org/abs/2203.02155}, 
}

@misc{shao2024deepseekmathpushinglimitsmathematical,
      title={DeepSeekMath: Pushing the Limits of Mathematical Reasoning in Open Language Models}, 
      author={Zhihong Shao and Peiyi Wang and Qihao Zhu and Runxin Xu and Junxiao Song and Xiao Bi and Haowei Zhang and Mingchuan Zhang and Y. K. Li and Y. Wu and Daya Guo},
      year={2024},
      eprint={2402.03300},
      archivePrefix={arXiv},
      primaryClass={cs.CL},
      url={https://arxiv.org/abs/2402.03300}, 
}

@misc{guo2025deepseek,
  title={Deepseek-r1: Incentivizing reasoning capability in llms via reinforcement learning},
  author={Guo, Daya and Yang, Dejian and Zhang, Haowei and Song, Junxiao and Wang, Peiyi and Zhu, Qihao and Xu, Runxin and Zhang, Ruoyu and Ma, Shirong and Bi, Xiao and others},
  journal={arXiv preprint arXiv:2501.12948},
  year={2025}
}

@misc{jin2025searchr1trainingllmsreason,
      title={Search-R1: Training LLMs to Reason and Leverage Search Engines with Reinforcement Learning}, 
      author={Bowen Jin and Hansi Zeng and Zhenrui Yue and Jinsung Yoon and Sercan Arik and Dong Wang and Hamed Zamani and Jiawei Han},
      year={2025},
      eprint={2503.09516},
      archivePrefix={arXiv},
      primaryClass={cs.CL},
      url={https://arxiv.org/abs/2503.09516}, 
}

@misc{xia2026search,
  title={Search-P1: Path-Centric Reward Shaping for Stable and Efficient Agentic RAG Training},
  author={Xia, Tianle and Xu, Ming and Hu, Lingxiang and Sun, Yiding and Li, Wenwei and Shang, Linfang and Liu, Liqun and Shu, Peng and Yu, Huan and Jiang, Jie},
  journal={arXiv preprint arXiv:2602.22576},
  year={2026}
}

@misc{wang2025ragenunderstandingselfevolutionllm,
      title={RAGEN: Understanding Self-Evolution in LLM Agents via Multi-Turn Reinforcement Learning}, 
      author={Zihan Wang and Kangrui Wang and Qineng Wang and Pingyue Zhang and Linjie Li and Zhengyuan Yang and Xing Jin and Kefan Yu and Minh Nhat Nguyen and Licheng Liu and Eli Gottlieb and Yiping Lu and Kyunghyun Cho and Jiajun Wu and Li Fei-Fei and Lijuan Wang and Yejin Choi and Manling Li},
      year={2025},
      eprint={2504.20073},
      archivePrefix={arXiv},
      primaryClass={cs.LG},
      url={https://arxiv.org/abs/2504.20073}, 
}

@misc{feng2025groupingrouppolicyoptimizationllm,
      title={Group-in-Group Policy Optimization for LLM Agent Training}, 
      author={Lang Feng and Zhenghai Xue and Tingcong Liu and Bo An},
      year={2025},
      eprint={2505.10978},
      archivePrefix={arXiv},
      primaryClass={cs.LG},
      url={https://arxiv.org/abs/2505.10978}, 
}

@article{mckenzie2023inverse,
  title={Inverse scaling: When bigger isn't better},
  author={McKenzie, Ian R and Lyzhov, Alexander and Pieler, Michael and Parrish, Alicia and Mueller, Aaron and Prabhu, Ameya and McLean, Euan and Kirtland, Aaron and Ross, Alexis and Liu, Alisa and others},
  journal={arXiv preprint arXiv:2306.09479},
  year={2023}
}
%




\end{document}